\documentclass[review,12pt,3p]{elsarticle}
\usepackage{framed,multirow}
\usepackage{amssymb}
\usepackage{latexsym}
\usepackage{url}
\usepackage{xcolor}
\usepackage{hyperref}
\usepackage{graphicx}
\usepackage{caption}
\usepackage{color}
\usepackage{adjustbox}
\usepackage{amsmath}
\usepackage{xspace}
\usepackage{booktabs}
\usepackage{setspace}
\usepackage{lineno}

\newcommand{\ie}{i.e.\@\xspace}
\newcommand{\ve}[1]{\mathbf{#1}}
 
\def\x{$\times$}

\definecolor{newcolor}{rgb}{.8,.349,.1}
\journal{Journal of Visual Communication and Image Representation}
\begin{document}
\begin{frontmatter}
\title{Salient Object Detection in Video using \\Deep Non-Local Neural Networks}

\author[]{Mohammad Shokri}
\ead{shokri@mail.um.ac.ir}
\author[]{Ahad Harati \corref{cor1}}
\ead{a.harati@um.ac.ir}
\author[]{Kimya Taba}
\address{Department of Computer Engineering, Ferdowsi University of Mashhad, Mashhad, Iran}
\cortext[cor1]{Corresponding author}

\begin{abstract}
  Detection of salient objects in image and video is of great importance in many computer vision applications. In spite of the fact that the state of the art in saliency detection for still images has been changed substantially over the last few years, there have been few improvements in video saliency detection. This paper investigates the use of recently introduced non-local neural networks in video salient object detection. Non-local neural networks are applied to capture global dependencies and hence determine the salient objects. The effect of non-local operations is studied separately on static and dynamic saliency detection in order to exploit both appearance and motion features. A novel deep non-local neural network architecture is introduced for video salient object detection and tested on two well-known datasets DAVIS and FBMS. The experimental results show that the proposed algorithm outperforms state-of-the-art video saliency detection methods.

\end{abstract}

\begin{keyword}
Video Saliency Detection \sep Deep Learning \sep Non-Local Neural Networks \sep Fully Convolutional Neural Networks
\end{keyword}
\end{frontmatter}

% \doublespace
% \linenumbers

\section{Introduction} \label{sec:introduction}
Humans are able to discern salient objects in their view effortlessly via pre-attentive processing. This helps a great deal in selecting the most informative parts of an image for further processing as it is the case in tasks such as visual scene segmentation where stimuli are grouped together into specific objects against a background \cite{Preattentive2009}. This capability has long been studied by cognitive scientists and attracted great amount of attention in computer vision community due to its effectiveness for finding important regions and objects in an image \cite{SalientObjectDetectionSurvey2014}. Saliency detection has been successfully applied to  many computer vision problems such as object detection \cite{SaliencyGuidedRegionProposal2017, ObjectDetectionSelectiveAttention2014, VOCUS2006, IntegratedAttentionOptimizingDetection2006}, image/video segmentation \cite{MultipleForegroundSegmentation2017}, image/video retrieval \cite{VideoHashingDBN2016}, video summarization \cite{VideoSummarizationAttention2017} and action recognition \cite{SaliencyAwareActionRecognition2017}.

Early attention models were focused on predicting human eye fixation \cite{SaliencyAttentionSceneAnalysis1998, SaliencySearchItti2000} whereas the recent methods have been more concentrated on detecting the salient object in a scene\cite{SalientObjectDetectionSurvey2014}. While these approaches both output a saliency map indicating the importance of each pixel in the input image and hence can be used interchangeably as a pre-processing step in computer vision applications, this paper is more inclined towards the latter. 

From a different perspective, saliency detection methods can be roughly divided into two different categories: bottom-up \cite{FrequencyTunedSRD2009, GlobalContrastSaliency2011, CenterSurroundedSaliency2011} and top-down approaches \cite{GenericObjectnessSaliency2011, ContextAwareSaliency2010, ObjectLevelSaliency2013}. The former methods are stimulus-driven, which infer the human attention based on visual stimuli themselves without the knowledge of the image content. In contrast, the top-down attention mechanisms are task-driven and usually require explicit understanding of the context of scene \cite{DeepAttentionPrediction2018}. Since the proposed method is believed not to depend on any prior knowledge about the context, it can be classified rather as a bottom-up approach.

Saliency detection has been an active research area in computer vision for a long time but most of the previous research has been focused on still-image saliency detection \cite{SaliencyAttentionSceneAnalysis1998, IntegratedAttentionOptimizingDetection2006, GlobalContrastSaliency2011, SaliencyFilters2012, UniversalSaliencyFramework2016, MultiTaskDeepSaliency2016, SpectralGraph2018}. During the last couple of years, saliency detection in videos has gained a lot of interest as well \cite{FeatureHybridFrameworkSaliency2018, ConsistentSaliencyGradientFlow2015, VideoSaliencyDeepFeatures2018, SalientRegionDetection2015, MultiScaleConvLSTMSaliency2018, ObjectToMotionLCNN2017, RegionBasedSaliency2017, SpatioTemporalDynamicSaliency2018, TimeMappingSaliency2014, VideoSaliency3dCNN2018, VideoSaliencyObjectProposals2018, VideoSaliencyDynamicAttention2013, VideoSaliencyCoherencyDiffusion2017, VideoSaliencyUncertaintyWeighting2014, BaggingBasedPrediction2018, VideoSaliencyDeepFeatures2018, VideoSaliancyFCN2018}. Most of these methods try to incorporate motion cues into the previously designed saliency detection models and use them together with appearance features to predict video saliency. Using motion features beside appearance features may be of invaluable help to provide indication for the visual foregrounds in videos, however the presence of background motions brings difficulties for locating moving objects; hence it might be challenging to exploit motion features in many cases. 

The re-emergence of convolutional neural networks (CNN) has brought about a significant improvement in a wide range of computer vision areas. The abundance of publicly available datasets \cite{ImageNet2009, COCO2014} together with the considerable advancements in GPU technology has made it possible to utilize very deep neural networks in real-time applications surpassing the accuracy of many state-of-the-art methods. In recent years, due to the successful deployment of deep neural networks in applications such as object detection \cite{FastRCNN2015, YOLO2016, SSD2016} and image/video segmentation \cite{OSVOS2017, SaliencyAwareObjectSegmentation2015, UNET2015}, these models have been the center of attention in almost every computer vision research area. Saliency detection was of course not an exception in this revolutionary path as there have been substantial improvements over the previous methods by utilizing the extraordinary discriminative power of CNNs \cite{UnconstrainedSaliencyProposal2016, SaliencyShortConnections2018, MultiTaskDeepSaliency2016, HirerchicalSaliency2016, SalientObjectRecurrentFCNN2018, UnconstrainedSalientObject2017}. Although these methods have been quite successful for saliency detection in image, directly applying them to video salient object detection is challenging due to the dramatic appearance contrast change and camera motion in videos \cite{FlowGuidedSaliency2018}.

The introduction of recent non-local neural networks \cite{NonLocal2018} which is believed to capture important long-range dependencies, has drawn lots of attention during the last couple of months \cite{MaskRCNN2017, AttentionUnet2018, Difnet2018, NonLocalRecurrentImageRestoration2018}. Non-local operations are introduced as a generic family of building blocks for capturing long-range dependencies and they can be employed in many existing models. Since it has been shown that they can extract global features which are not captured by conventional models through repetition of convolutional operations, we investigate their application in video salient object detection. The importance of global dependencies in determining salient parts of an image or a video, motivated us to scrutinize the potential influence of applying non-local neural networks for video salient object detection.

We evaluate the proposed method on two well-known datasets: DAVIS \cite{DAVIS2016} and FBMS \cite{FBMS2014} and show the effectiveness of using non-local neural networks for improving the accuracy of state-of-the-art methods. We also report the performance and time-efficiency of the proposed method which makes it feasible to be applied to real-time applications. 

\section{Related works}

\subsection{Still-image saliency detection}
Detection of salient regions or objects in still-images has been widely studied during the last two decades. Image saliency detection aims at discovering the most distinctive object in an individual image through some visual priors and contrast in appearance. Early approaches were mostly trying to mimic the human attention mechanisms inspired by the studies on human visual system; therefore they were focused on predicting human eye fixations however in the latest saliency detection studies, identifying the salient regions from the image was of more concern \cite{VisualSaliencyReview2018}. Conventional methods were performed either via bottom-up approaches based on low-level features or in a top-down fashion through the incorporation of high-level knowledge considering the target object in the image. Similar to other machine vision fields, deep CNN has led salient object detection into a new frontier and helped researchers to significantly improve the state of the art in this field. Deep CNN-based models for saliency detection can be classified into two main categories, including region-based learning methods \cite{VisualSaliencyMultiscale2015, DeepSaliencyLocalEstimation2015, RegionBasedSaliency2017} and end-to-end fully convolutional networks \cite{InstanceLevelSaliency2017,DeepContrastSalientObject2016, MultiTaskDeepSaliency2016}. Deep fully convolutional based methods have the advantage of feature sharing over region-based methods and are generally faster \cite{FlowGuidedSaliency2018}. In \cite{UnconstrainedSalientObject2017} saliency map prediction is done without pixel-level annotation through modeling salient region as a Gaussian distribution while training a CNN. Recurrent FCNs are also applied for predicting saliency in order to automatically learn to refine generated saliency map by correcting previous errors \cite{SalientObjectRecurrentFCNN2018}. Deep CNN-based models have obtained superior results over the past few years and become an essential part of any state-of-the-art method in image saliency detection.

\subsection{Video saliency detection}
In contrast to the aforementioned saliency prediction methods for still-images, video saliency detection methods can benefit from temporal and motion information which exists in video streams. This information can be integrated in previous FCN-based methods and extend them for the task of video salient object detection. Nevertheless benefiting from motion features and combining them with image saliency models requires new network architectures. Only recently have such models been proposed and applied to video salient object detection task \cite{FlowGuidedSaliency2018, VideoSaliancyFCN2018}. Prior to the re-advent of CNNs, motion features were employed in video saliency detection through keypoint correspondence between frames using homography \cite{SpatiotemporalAttention2006}, motion fields computed by optical flow \cite{VideoAttention2008} and temporal coherence based energy minimization framework \cite{SegmentingSalientObject2010}. Considering the noticeable impact of moving objects in attracting human attention, an object-to-motion convolutional neural network (OM-CNN) is presented in \cite{ObjectToMotionLCNN2017} for estimating intra-frame saliency through leveraging the information of both objectness and object-motion. Object proposals have also been utilized for saliency detection by ranking and selecting the salient proposals according to the saliency cues and then combining the results with motion contrast-based saliency \cite{VideoSaliencyObjectProposals2018}.

In an effort to combine spatial and temporal information with the aim of predicting salient regions in video, the authors of \cite{VideoSaliencyUncertaintyWeighting2014} have proposed an adaptive entropy-based uncertainty weighting approach which take into account proximity and continuity of spatial saliency along with variations of background motion. A two-stream deep network architecture is suggested in \cite{SpatioTemporalDynamicSaliency2018} for integrating spatial and temporal information to predict saliency maps by exploiting optical flow maps. In another attempt to utilize local and global contexts over frames, a new set of spatiotemporal deep features (STD) have been proposed in \cite{VideoSaliencyDeepFeatures2018} to be used for detecting salient objects. Recurrent neural networks (RNNs) have also been studied for video saliency detection via incorporating spatial and temporal cues in multi-scale spatiotemporal convolutional LSTM network \cite{MultiScaleConvLSTMSaliency2018} and flow guided recurrent neural encoder to compensate camera motion and dramatic change of appearance contrast in videos \cite{FlowGuidedSaliency2018}. Spatiotemporal features in video sequences are learned using a 3d convolutional network in \cite{VideoSaliency3dCNN2018} with three video frames as input of the network. A stack of convolutional layers similar to \cite{VeryDeepCNN2014} followed by transposed convolutional layers are used in \cite{VideoSaliancyFCN2018} to predict static saliency and then a similar network architecture is designed to predict dynamic saliency taking two consecutive video frames along with static saliency as prior. We have used this network architecture as a baseline since it is a fast method which produces promising results and extended it with non-local operations to achieve superior results while barely increasing its computational load.

\section{Proposed approach}

Both convolutional and recurrent operations are supposed to take local neighborhood information into account either in space or time dimension. Long-range dependencies and global information is expected to be obtained through applying repeated local operations and thus propagating signals through space or time. Apart from computational inefficiency of such repeated operations to extract global information, it is suggested in \cite{NonLocal2018} that repeated conventional operations are unable to discover all existing global dependencies through comprehensive experience. The additional global information that can be obtained by applying just a few non-local blocks and hence their computational cost-effectiveness is of invaluable benefit for salient object detection. Since non-local features are inherently relevant to saliency both in space and time, the effect of exploiting non-local features is studied separately for static and dynamic saliency prediction. We extend the architecture proposed in \cite{VideoSaliancyFCN2018} by exploiting the extra information extracted via adding non-local blocks to the fully convolutional neural networks for video salient object detection.

The output of both static and dynamic saliency networks is a pixel-wise probability map indicating the probability of each pixel being a part of the salient object in the input image, \ie brighter pixels in the output map show more salient parts of the image. The two steps of predicting saliency values are applied sequentially for each frame of the video.

\subsection{Non-local saliency block}

\begin{figure}[h]
  \centering
  \includegraphics[width=\linewidth]{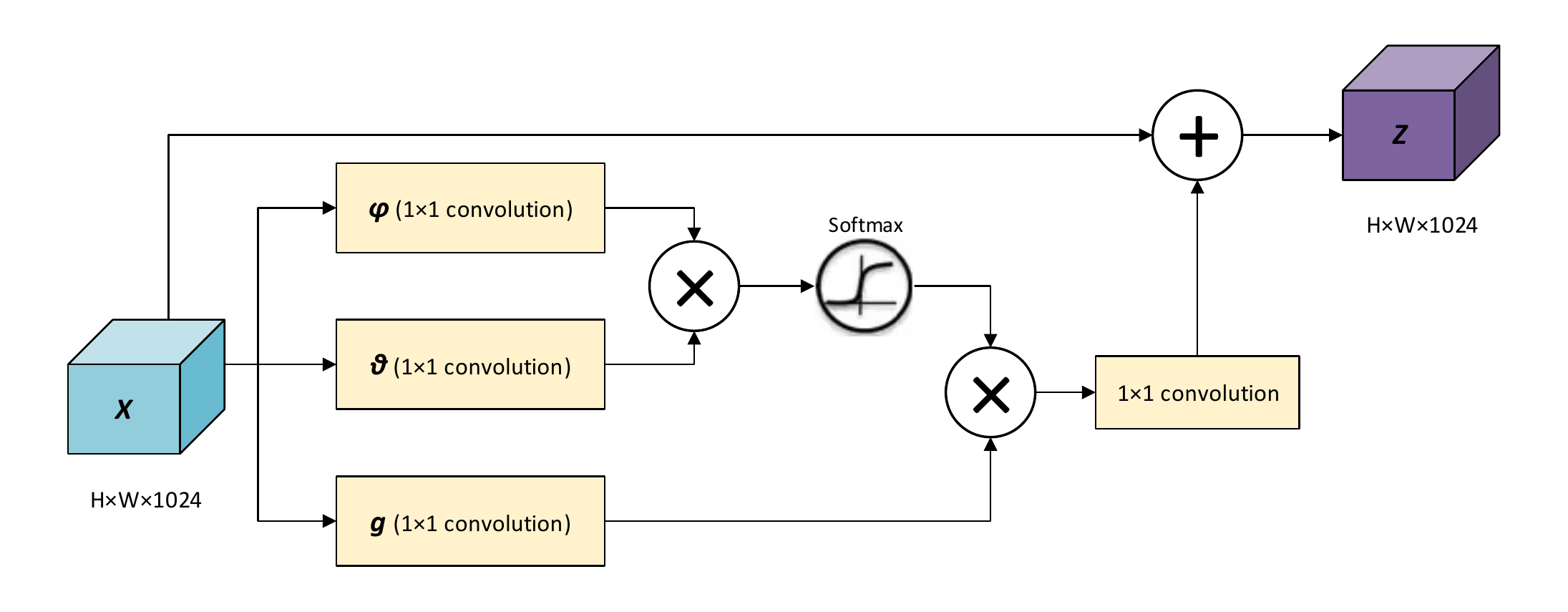}
  \caption{\textbf{Non-local block}. The input feature map $\ve{x}$ is fed into the block as a tensor of size $H$\x$W$\x$1024$ which is the same size as the output of the whole block. This is a spatial extension of the general non-local block introduced in \cite{NonLocal2018}. ``$\otimes$'' denotes matrix multiplication, and ``$\oplus$'' denotes element-wise sum and the softmax operation is performed on each row.}
  \label{fig:nl-block}
\end{figure}

In order to exploit more global information for detecting salient object, non-local operations are employed in both static and dynamic saliency prediction networks in quite the same way. Inspired by the original non-local neural networks, the non-local saliency operation is defined as:
\begin{equation}\label{eq:nonlocal}
  \ve{y}_{i,j} = \frac{1}{C(\ve{x})} \sum_{\forall {k,l}}f(\ve{x}_{i,j}, \ve{x}_{k,l})g(\ve{x}_{k,l}).
\end{equation}
where $(i,j)$ are the coordinates of the position whose response is to be computed and $(k,l)$ are the coordinates of all possible positions in the input image. $\ve{x}$ is the input image or its correspondent feature map and $\ve{y}$ is the output signal of the same size as $\ve{x}$. The pairwise function $f$ computes a scalar between $(i,j)$  and all $(k,l)$ and $g$ is a unary function of the input signal at the position $(k,l)$. The output values are normalized by a factor ${C}(\ve{x})$.

The unary function $g$ is computed through a linear embedding: 
\begin{equation}\label{eq:nonlocal_g}
  g(\ve{x}_j)=W_g\ve{x}_j.
\end{equation}
where $W_g$ is a weight matrix to be learned. For the implementation of $g$, a 1\x1 two-dimensional convolution is used.

Here $f$ is assumed to be embedded Gaussian function, since it is shown to perform better in our experiments and therefore it is defined as follows:
\begin{equation}\label{eq:nonlocal_f}
  f(\ve{x}_{i,j}, \ve{x}_{k,l}) = e^{\theta(\ve{x}_{i,j})^T\phi(\ve{x}_{k,l})}.
\end{equation}
where $\theta(\ve{x}_{i,j})=W_\theta\ve{x}_{i,j}$ and $\phi(\ve{x}_{k,l})=W_\phi\ve{x}_{k,l}$ are two embeddings which are implemented using two-dimensional convolutions and $\mathcal{C}(\ve{x})$ is calculated via:
\begin{equation}\label{eq:nonlocal_c}
\begin{aligned}
  \mathcal{C}(\ve{x})=\sum_{\forall {k,l}}f(\ve{x}_{i,j}, \ve{x}_{k,l}).
  % \theta(\ve{x}_{i,j})=W_\theta\ve{x}_{i,j}, \\
  % \phi(\ve{x}_{k,l})=W_\phi\ve{x}_{k,l},
\end{aligned}
\end{equation}
since for a given position $(i,j)$, $\frac{1}{{C}(\ve{x})}f(\ve{x}_{i,j}, \ve{x}_{k,l})$ becomes the \emph{softmax} computation along the dimensions $(k,l)$, values of the output signal $\ve{y}$ can be calculated using $\ve{y}={\emph{softmax}}(\ve{x}^TW^T_\theta W_\phi\ve{x})g(\ve{x})$.

In order for the model to be able to use any pre-trained model in previous layers, the final output of the whole non-local block is defined as:
\begin{equation}\label{eq:block}
\ve{z}_{i,j} = W_z \ve{y}_{i,j} + \ve{x}_{i,j},
\end{equation}
where $\ve{y}_{i,j}$ is given in Eq.(\ref{eq:nonlocal}) and ``$+ \ve{x}_{i,j}$'' denotes a residual connection. The residual connection allows us to insert a new non-local block into any pre-trained model.
All of the two-dimensional convolutions mentioned above, use 1\x1 kernels, 1\x1 strides, \emph{same} padding and \emph{ReLU} activation function.
The non-local block used in the following sections is depicted in Figure~\ref{fig:nl-block}.

\subsection{Static saliency detection}

\label{sec:static_saliency}
\begin{figure}[t]
  \centering
  \includegraphics[width=\linewidth]{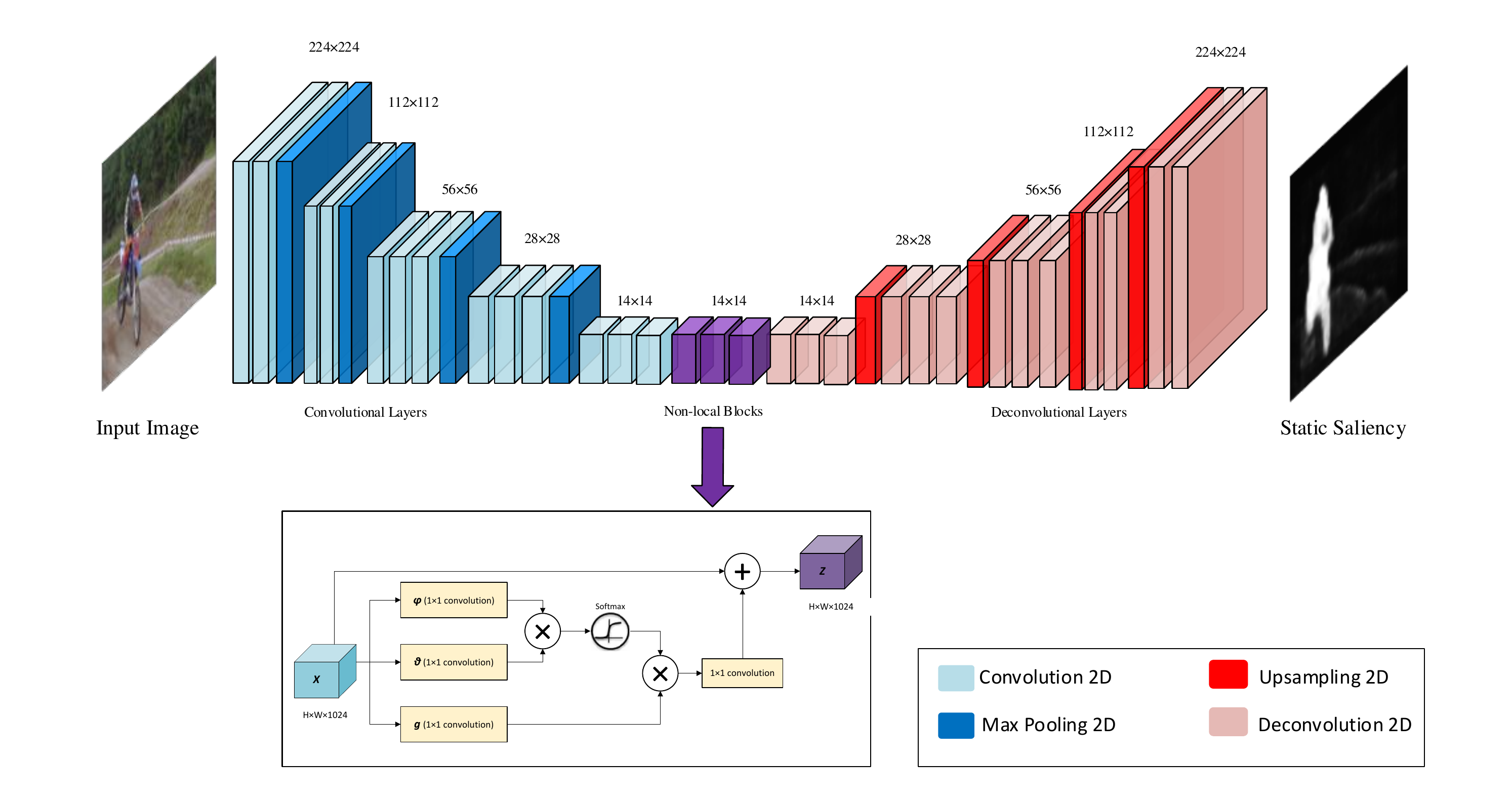}
  \caption{\textbf{Static saliency detection.} Convolutional layers extract features from the input frame image, followed by 3 non-local blocks to add more long-range features to the extracted features. The output of non-local blocks is then upsampled multiple times through deconvolutional part of the network and then a pixel-level saliency probability map is estimated applying a fully convolutional network with 1\x1 kernel size and \emph{sigmoid} activation function at the end. }
  \label{fig:static-saliency}
\end{figure}

As previously stated, salient object in the video is predicted in two sequential steps namely static and dynamic saliency detection. The former employs only one frame of the video to predict its saliency map taking only appearance features into account, while the latter is expected to correct predictions of the first stage regarding motion saliency predicted using two consecutive frames as input. In order for the one-frame static saliency to be predicted, a deep CNN-based model is applied to each frame of the video. A general overview of the proposed architecture for static saliency detection is illustrated in Figure~\ref{fig:static-saliency}.
As suggested in \cite{VeryDeepCNN2014} in order to extract features from the input image, there are five blocks of VGG network in the first part of the network, each of which consists of convolution layers followed by max pooling layers. Each convolution layer uses a nonlinear function (ReLU) that is a perfect fit for the saliency prediction network regarding its computationally-efficiency and sparsity.
Following these features extraction layers, 3 non-local blocks are used to exploit long-range features that are fed to the subsequent deconvolution (or transposed convolution) layers which in turn are aimed to produce pixel-wise saliency map with a higher resolution.

The output of convolution layers preceding non-local blocks is computed as:
\begin{equation}
  \begin{aligned}
  Y = F(X;W,b) = W \ast X + b,
  \end{aligned}
\end{equation}
where $X$ is the input feature map and $W$ and $b$ are weights and bias used in convolution operation. Given the value of $Y$ as input of the non-local block $b$, the output of the block $b$ for input channel $c$ is as:

\begin{equation}
  \begin{aligned}
    Z_c={\emph{softmax}}(Y^TW^T_\theta W_\phi Y)W_g Y
  \end{aligned}
\end{equation}
where $W_\theta$, $W_\phi$ and $W_g$ are kernel weights to be learned. The output of the last non-local block is then given to the subsequent deconvolution layer as input resulting in the value:
\begin{equation}
  \begin{aligned}
    O_5=D(Z; \theta_5)
  \end{aligned}
\end{equation}
where $D$ denotes the deconvolution layers that are used to upsample input for obtaining an output with the same size of the input image. Each layer $l$ of the next 4 deconvolution layers takes the output of its preceding layer $l+1$ together with the output of the corresponding convolution layer $l$ as input and computes $O_l$ as:
\begin{equation}
  \begin{aligned}
    O_l=D((O_{l+1},Y_l); \theta_l)
  \end{aligned}
\end{equation}
where $(O_{l+1},Y_l)$ is the concatenation of $O_{l+1}$ and $Y_l$. At the end of the static saliency network, a convolutional layer with kernel size 1\x1 and a \emph{sigmoid} activation function is applied to predict saliency map $S$ indicating the saliency of each pixel using the values of the last deconvolution layer.

Considering the computational load of non-local operations when they are applied after different convolution layers, they have been used after the fifth block of the network as it has been observed that the advantage of using them in previous layers is insubstantial. The effect of inserting non-local blocks after different blocks of feature extraction network is studied and reported in section~\ref{sec:results}.

\subsection{Dynamic saliency detection}

\begin{figure}[t]
  \centering
  \captionsetup{justification=centering}  
  \includegraphics[width=\linewidth]{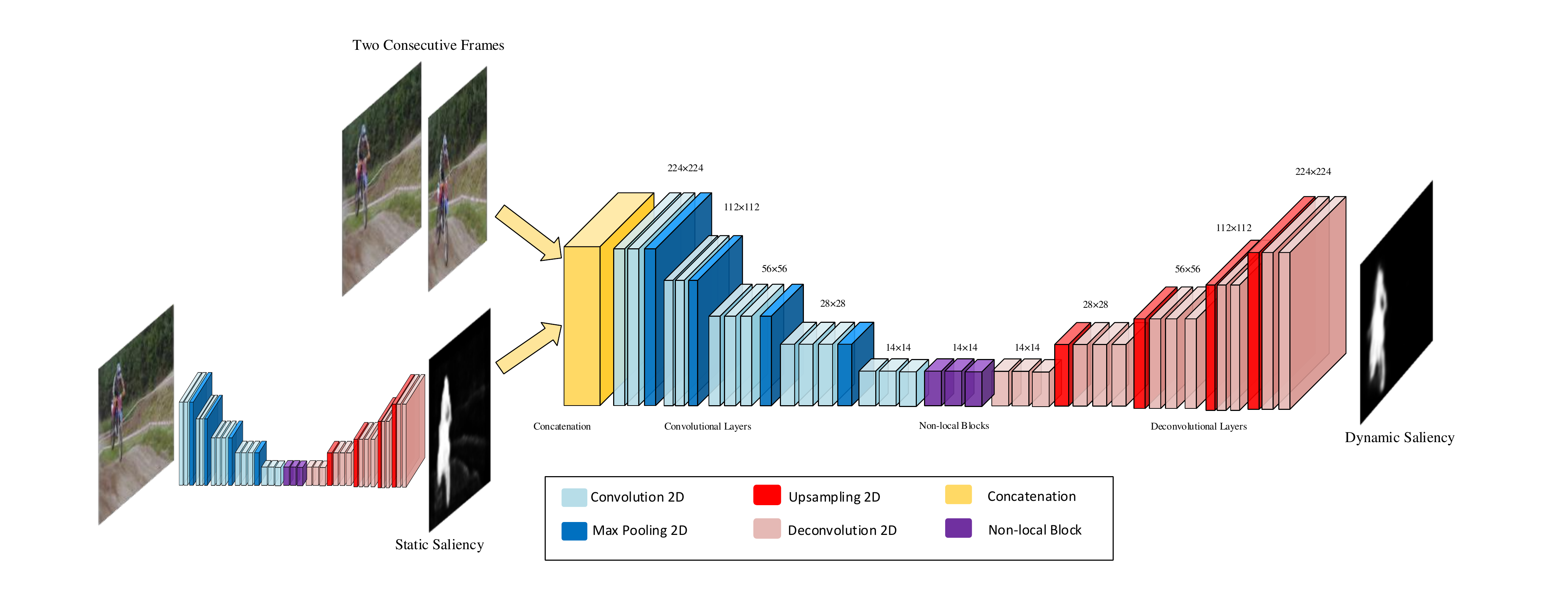}
  \caption{\textbf{Dynamic saliency detection.} Two consecutive video frames and the saliency predictions from the static saliency network are concatenated and used as input of the dynamic saliency prediction network.}
  \label{fig:dynamic-saliency}
\end{figure}

Employing motion features in a video can be of tremendous importance when detecting salient objects. In order to use motion information to correct predictions obtained from the static saliency network, two consecutive frames are given to a deep neural network as input alongside with results of the first step. It is expected that providing consecutive frames to the network, exploit motion features and enhance the predicted static saliency map. As illustrated in Figure~\ref{fig:dynamic-saliency}, the architecture of the deep network used in the second step is similar to the one mentioned in Section~\ref{sec:static_saliency} for the first step, except that the single frame input is replaced by two consecutive frames $(I_t, I_{t+1})$ plus the output of the static saliency detection $S_t$. The two RGB frames of the video and the static saliency map are concatenated through channels dimension to form the input of the dynamic saliency detection network as an input tensor with dimensions $h$\x$w$\x$7$. The rest of the layers used in convolution and deconvolution parts of the network remain unchanged. 

Using non-local blocks in dynamic saliency detection network is studied separately as it has different advantages than that of static saliency model. Apart from the benefits discussed in the previous sections, applying non-local blocks on motion features tends to distinguish between global movements and object motions, \ie camera motions and objects motions are expected to be treated differently. Therefore it can be regarded as an indirect motion compensation method for attending to the salient moving objects.

To train both static and dynamic saliency detection models, a cross-entropy loss function is employed. for any training sample $(I_t, G_t)$ consisting
of the image $I_t$ with size $h\times w \times 3$, and the groundtruth saliency map $G_t\in\{0, 1\}^{h\times w}$, the network outputs saliency map $S_t \in [0, 1]^{h\times w}$ at time $t$. For any given training sample, the
loss value for the predicted saliency map $S_t$ is thus computed as:
\begin{equation}
  \begin{aligned}
  \mathcal{L}(S_t,G_t) = -\sum_{i=1}^{h} \sum_{j=1}^{w} \big( g_{i,j} \log s_{i,j} + (1-g_{i,j}) \log (1-s_{i,j})\big),
  \end{aligned}
  \label{eq:3}
\end{equation}
where $g_{i,j}\in G_t$ and $s_{i,j}\in S_t$.

\section{Experimental results}
\label{sec:results}

\subsection{Implementation details}

We use Tensorflow \cite{tensorflow2015-whitepaper} framework throughout training and testing the proposed network architecture on Ubuntu 16.04 operating system using a PC equipped with a 3.4GHz Intel CPU, 64 GB RAM and an NVIDIA Titan X GPU. Since the proposed method can be integrated into any CNN based salient object detection method, a recently introduced method named fully convolutional network (FCN) is used as baseline method which is considerably fast. First five convolutional blocks of VGG model \cite{VeryDeepCNN2014} are transferred to the static saliency detection network and the rest of layers are trained using training samples from datasets in Section~\ref{sec:datasets}. All of the components incorporated in our saliency detection framework are trained using stochastic gradient descent (SGD) with a momentum of 0.9 and the loss function for both static and dynamic saliency detection networks is set to cross entropy loss. 

\subsection{Datasets}
\label{sec:datasets}

\begin{table}[t]
  \centering \caption{The specification of two publicly available datasets: DAVIS \cite{DAVIS2016} and FBMS \cite{FBMS2014} which are used in the experiments of this paper.}
  \begin{tabular}{  c | c | c | c }
  \hline Datasets&\#Clips&\#Frames&\#Annotations\\
  \hline\hline DAVIS \cite{DAVIS2016}&50&3455&3455\\
  \hline\hline FBMS \cite{FBMS2014}&59&13860&720\\  
  \hline\end{tabular}
  \label{datasets}
\end{table}

The proposed method is trained and evaluated together with some state-of-the-art methods on two well-known publicly available datasets DAVIS \cite{DAVIS2016} and FBMS \cite{FBMS2014}. The overall statistical information about these datasets is provided in table~\ref{datasets}.

\textbf{DAVIS} (Densely Annotated VIdeo Segmentation) dataset \cite{DAVIS2016} consists of 50 high quality video sequences released as Full HD images and also $480p$ frames which are used in all of our experiments due to computational concerns. All of the video frames in this dataset are annotated as binary images. It contains a total of 3455 frames in different visual contents including sports, vehicles, animals and outdoor scenes. Various challenges and attributes such as appearance change, occlusion, multi moving objects, camera motion and dynamic background are covered in these videos. The goal in this dataset is to segment the most salient or dominant object and only one salient object is present in every video sequence.

\textbf{FBMS-59} (Freiburg-Berkeley Motion Segmentation) dataset \cite{FBMS2014} contains 59 video sequences split into training and testing sets with 29 and 30 video sequences respectively. It includes traffic scenes, several short sequences of the series Mrs. Marple, some indoor scenes of cats and rabbits in addition to various outdoor scenes. There are 720 annotated frames in FBMS dataset which are used for training and evaluating in our experiments. The primary goal of this dataset is to segment moving objects in presence of typical challenges which are covered in videos. The moving objects in FBMS are annotated as different objects in groundtruth images. Since the differentiation between separate salient objects is irrelevant for us, all of the foreground pixels in groundtruth images are changed to a single value to separate only salient and non-salient pixels.

\subsection{Evaluation criteria}
\label{sec:criteria}

Following \cite{SalientObjectDetectionBenchmark2015}, different metrics are utilized for quantitative performance evaluations including precision and recall curve (PR curve), F-Measure, mean absolute error (MAE), receiver operating characteristics (ROC) curve and area under ROC curve (AUC).

In order to form a PR curve, each saliency map $S$ is converted to a binary map $M$ and the values of precision and recall regarding a ground truth $G$ are computed as: 
\begin{equation}
  \text{Precision} = \frac{|M\cap G|}{|M|}, \ \ \text{Recall} = \frac{|M\cap G|}{|G|}
\end{equation}
To compute a binary map $M$ using a predicted saliency map, we have used a fixed threshold which varies from 0 to 255. For each value of this threshold, a pair of precision/recall is then computed to create the final PR curve illustrating the object retrieval performance of the model at different situations. The balanced degree of object retrieval between precision and recall values is also computed as F-measure using the following equation: 
\begin{equation}
  \text{F-measure} = \frac{(1+\beta^{2} ) \times  Precision  \times  Recall}{\beta^{2}  \times  Precision + Recall}
  \label{eq:fmeasure}
\end{equation}
As suggested in \cite{SalientObjectDetectionBenchmark2015}, $\beta^{2}$ is set to $0.3$ to increase the importance of the $Precision$ value.

The ROC curves are plotted using false positive rates (FPR) and true positive rates (TPR) which are defined as:
\begin{equation}
  \text{TPR} = \frac{|M\cap G|}{|G|}, \ \
  \text{FPR} = \frac{|M\cap \bar{G}|}{|\bar{G}|}
\end{equation}
where $M$ and $G$ stand for the the binary map and groundtruth respectively and $\bar{G}$ indicates the complement of $G$. To plot the ROC curve, the values of $\text{TPR}$ and $\text{FPR}$ can be computed at different thresholds in the same way as forming the PR curve.

Since the number of true negative saliency assignments, \ie pixels which are correctly marked as non-salient, is not regarded in other criteria, the mean absolute error (MAE) is also used for evaluation. Provided that the values of saliency map $S$ and the groundtruth $G$ are normalized in the range of [0, 1], MAE can be computed as follows:
\begin{equation}
  MAE = \frac{1}{W \times H} \sum\nolimits_{i=1}^{W}\sum\nolimits_{j=1}^{H} | {S}(i,j) - {G}(i,j)|
  \label{MAE}
\end{equation}
where $W$ and $H$ denote the width and height of the input image.

\subsection{Performance comparison}

\begin{figure*}[!ht]
  \centering
  \begin{minipage}[t]{\linewidth} 
    \includegraphics[width=\linewidth]{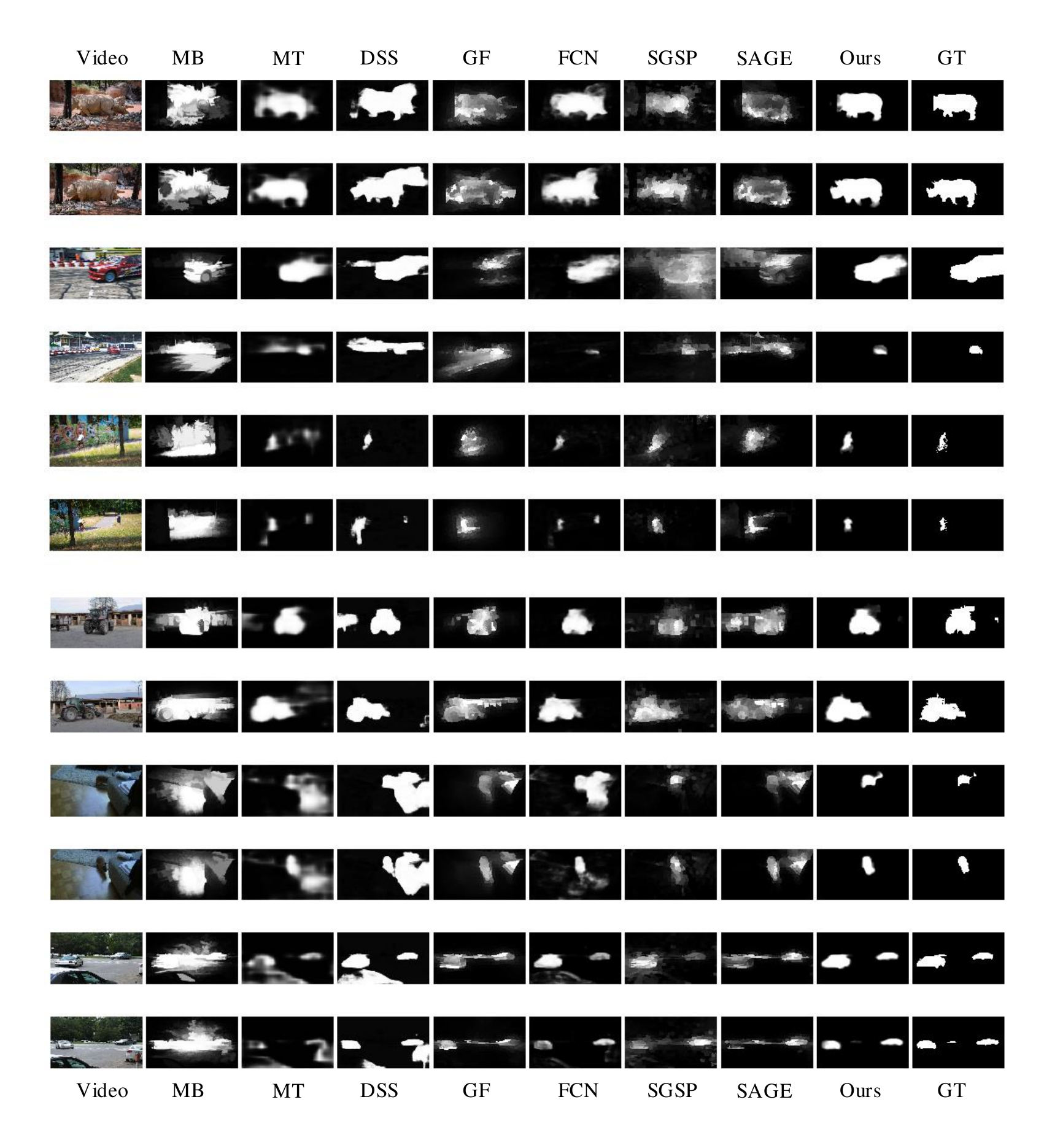}
    \caption{Saliency maps generated using different methods. Each two consecutive rows are from the same video sequence. Frames in top and bottom 6 rows are chosen from DAVIS and FBMS datasets, respectively.} 
    \label{fig:saliency-results}
  \end{minipage}
\end{figure*}

The proposed method is evaluated on both DAVIS and FBMS datasets and the results are compared against seven recent state-of-the-art saliency detection methods: deeply supervised salient object detection (\textbf{DSS}) \cite{SaliencyShortConnections2018}, minimum barrier (\textbf{MB}) \cite{MinimumBarrierSaliency2015}, multi-task deep neural network (\textbf{MT}) \cite{MultiTaskDeepSaliency2016}, local gradient flow optimization (\textbf{GF}) \cite{ConsistentSaliencyGradientFlow2015}, superpixel-level graph (\textbf{SGSP}) \cite{UnconstrainedSuperpixelSaliency2017}, geodesic distance based video saliency (\textbf{SAGE})\cite{SaliencyAwareObjectSegmentation2018} and fully convolutional networks (\textbf{FCN}) \cite{VideoSaliancyFCN2018}. The first three are still-image saliency detection methods while the last four methods operate on video sequences to predict saliency maps. Either saliency maps or the implementations provided by the authors are used for evaluation of all methods. 

\subsubsection{Qualitative performance comparison}
The predicted saliency maps for several frames of DAVIS and FBMS datasets, using different methods are visualized in Figure~\ref{fig:saliency-results}. Brighter pixels in the output saliency maps indicate more salient regions. It can be seen from the sample qualitative results that the proposed method is able to make significant improvements over the baseline method (FCN) using non-local information to predict saliency maps. A side effect of more attention to non-local information is the reduced sharpness which can be observed around the boundaries of the salient object detected by the proposed method. This can be avoided through employing non-local blocks after earlier convolutional blocks with higher resolution inputs, though axiomatically this improvement comes at the cost of more computational time.

As it can be observed in predicted saliency maps in Figure~\ref{fig:saliency-results}, still-image saliency detection methods face difficulties predicting the salient object in video sequences, yet deep learning based image saliency detection methods (DSS and MT) seem to handle video frames better than other static saliency detection methods.

\subsubsection{Quantitative performance comparison}

\begin{figure*}[!ht]
  % \centering
  \begin{minipage}[b]{0.3\linewidth} 
    \includegraphics[width=\linewidth]{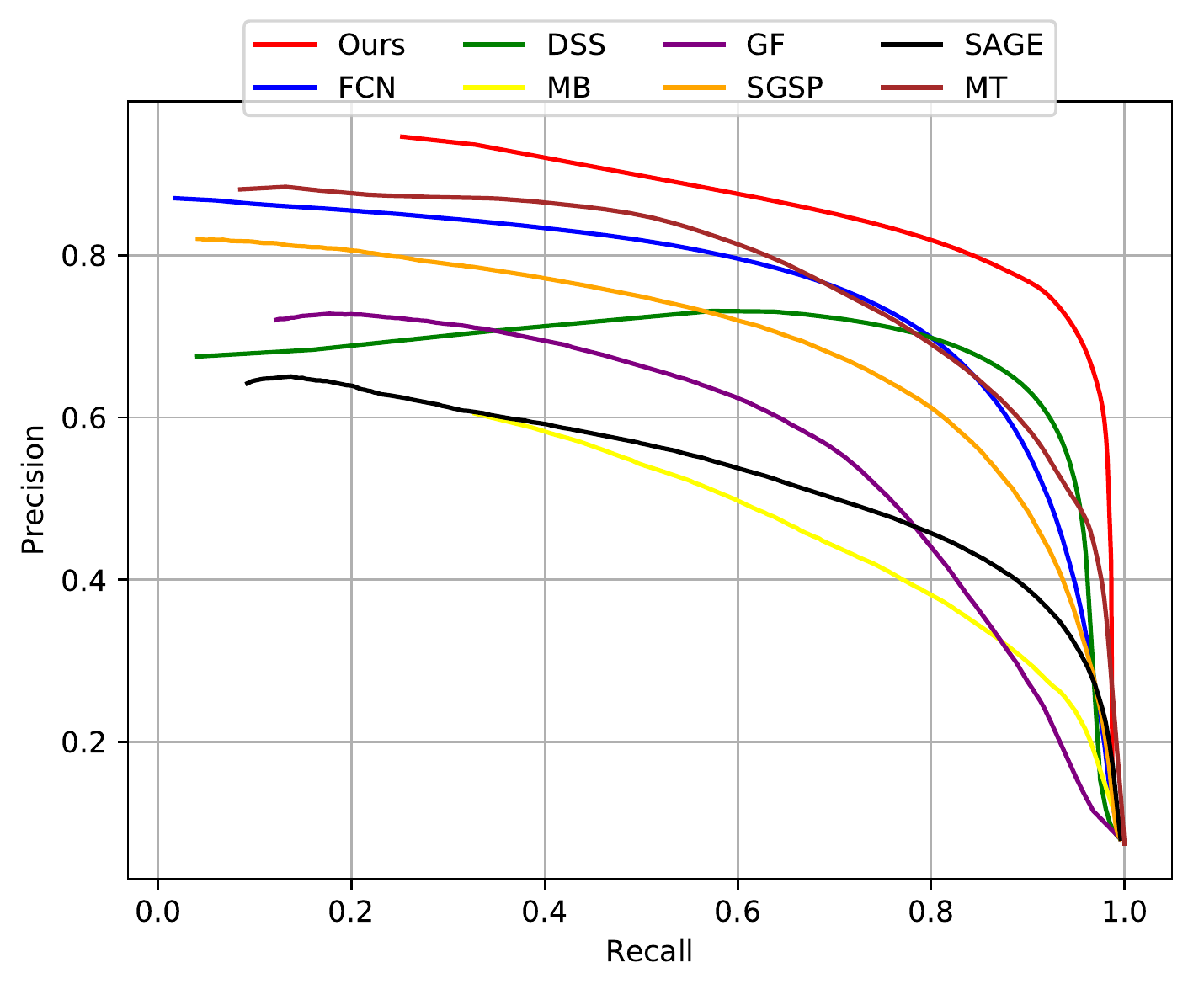}
  \end{minipage}
  \begin{minipage}[b]{0.3\linewidth}
    \includegraphics[width=\linewidth]{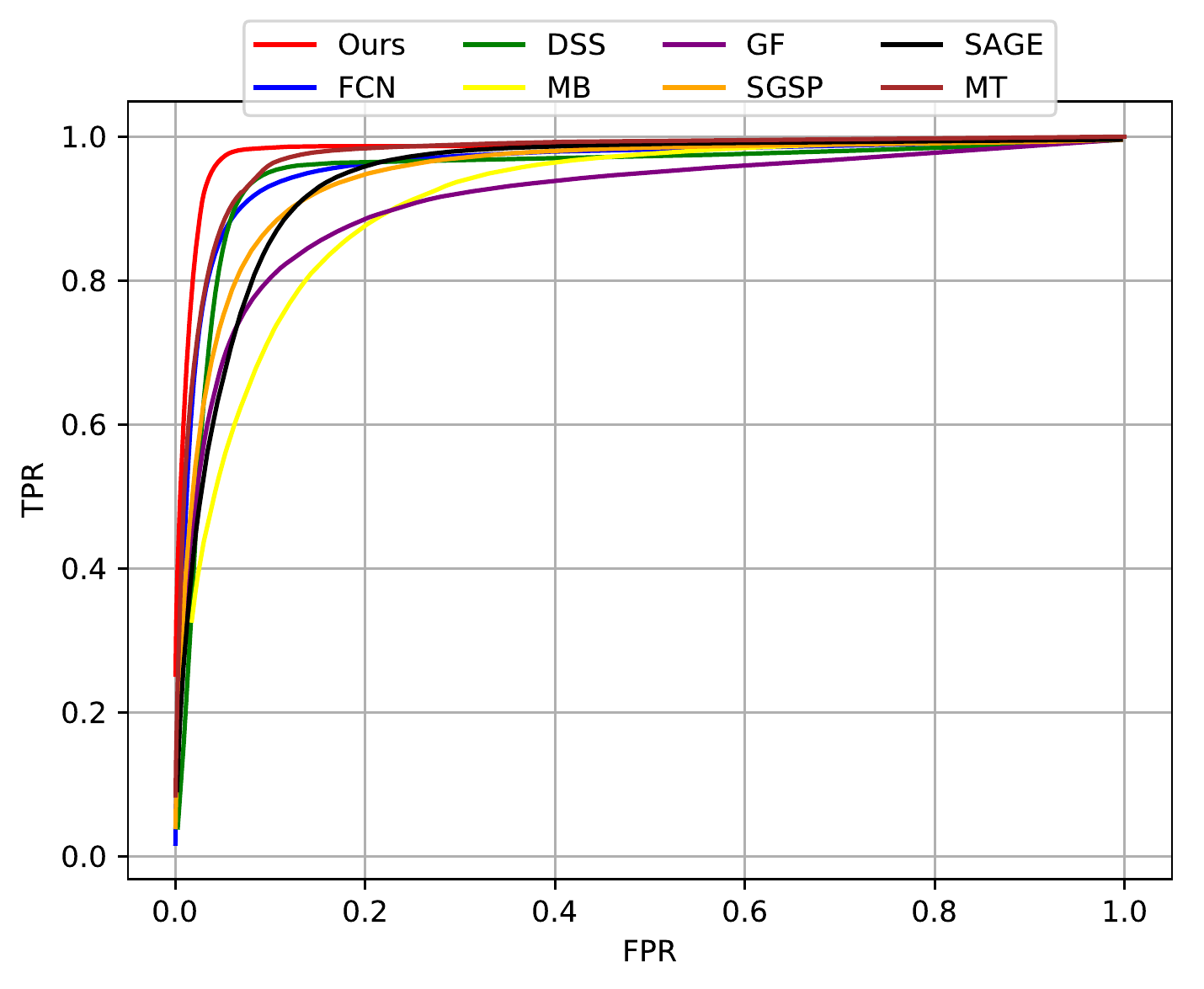}
  \end{minipage}
  \begin{minipage}[b]{0.3\linewidth}
    \includegraphics[width=\linewidth]{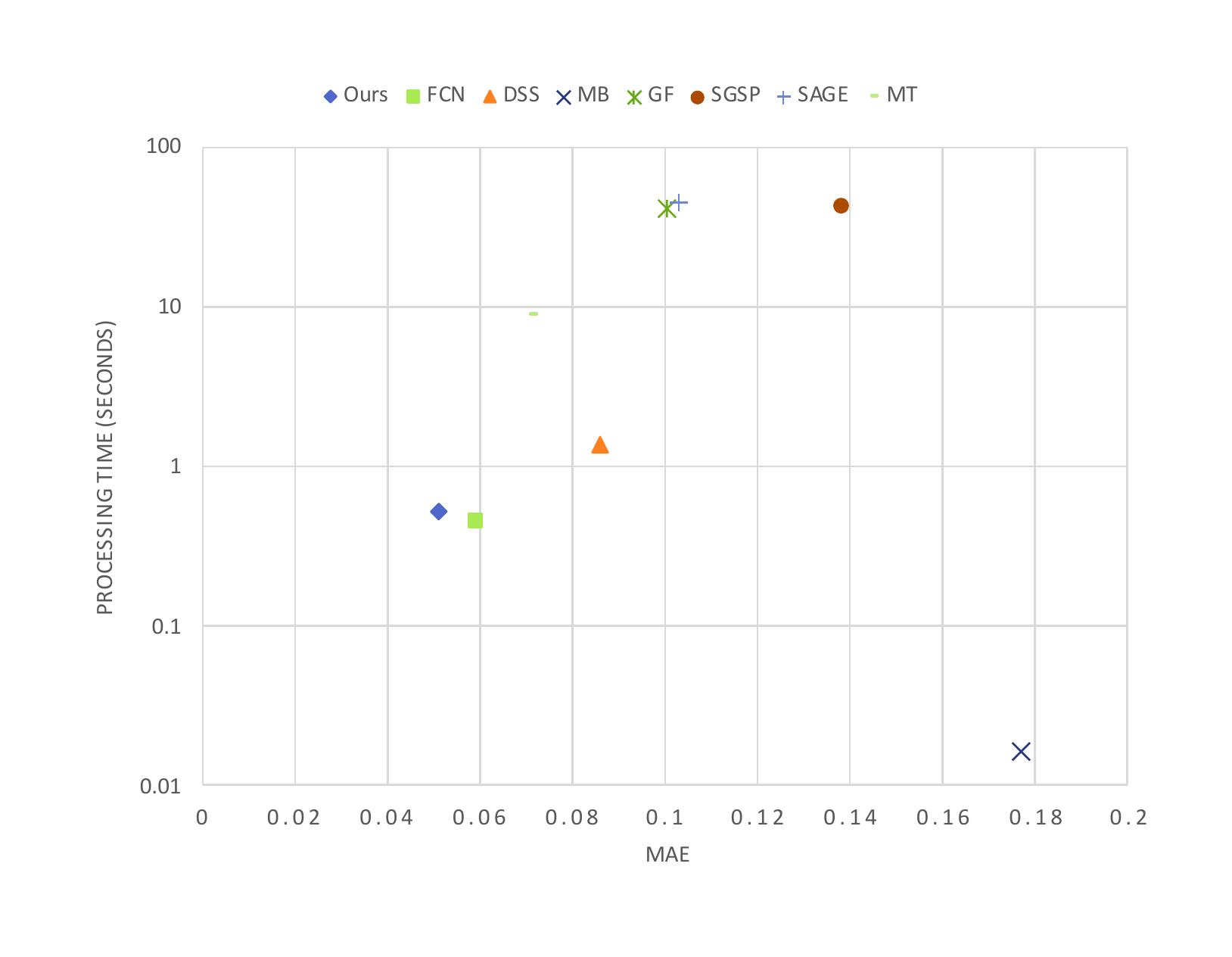}
  \end{minipage}

  \begin{minipage}[b]{0.3\linewidth} 
    \includegraphics[width=\linewidth]{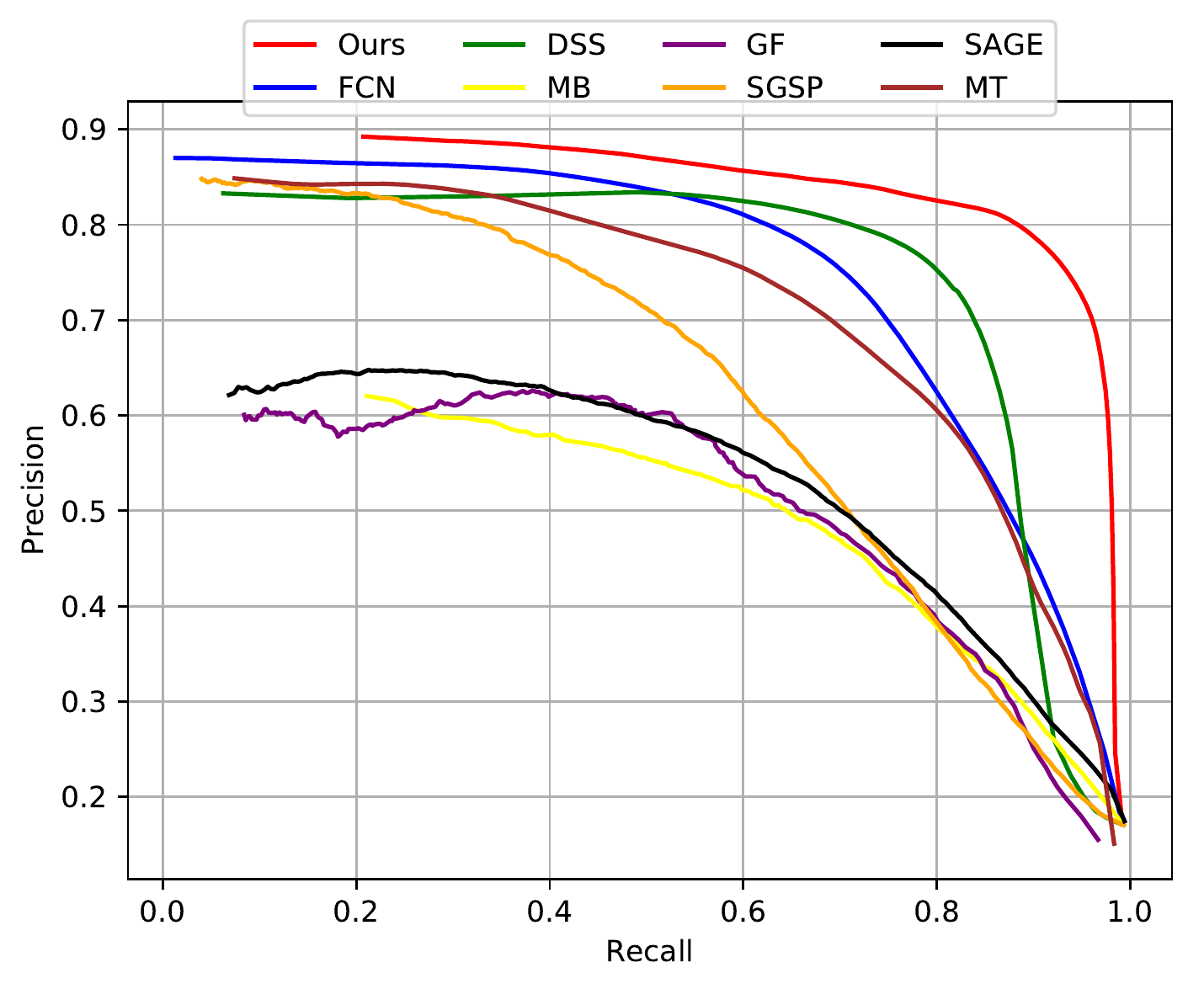}
  \end{minipage}
  \begin{minipage}[b]{0.3\linewidth}
    \includegraphics[width=\linewidth]{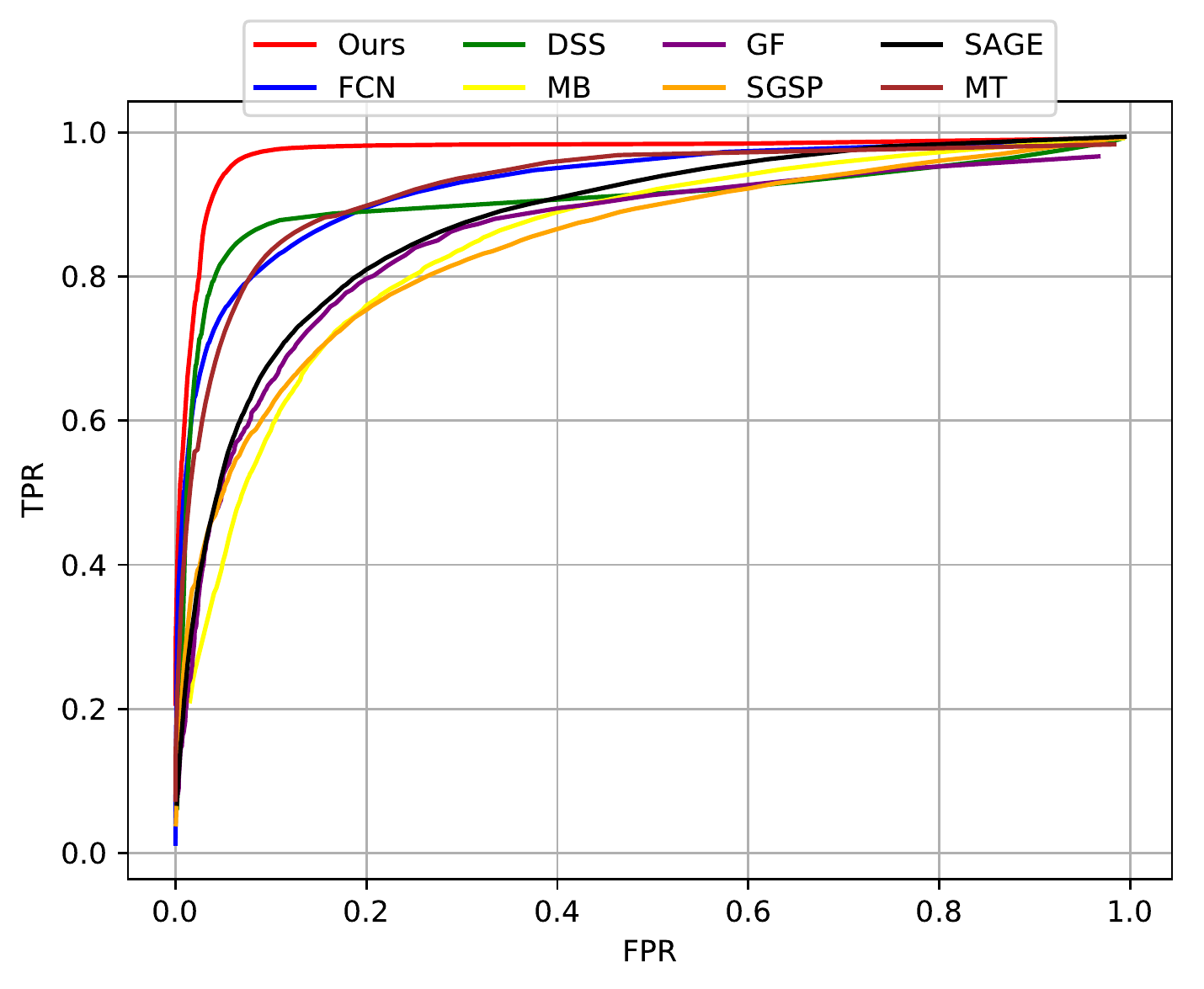}
  \end{minipage}
  \begin{minipage}[b]{0.3\linewidth}
    \includegraphics[width=\linewidth]{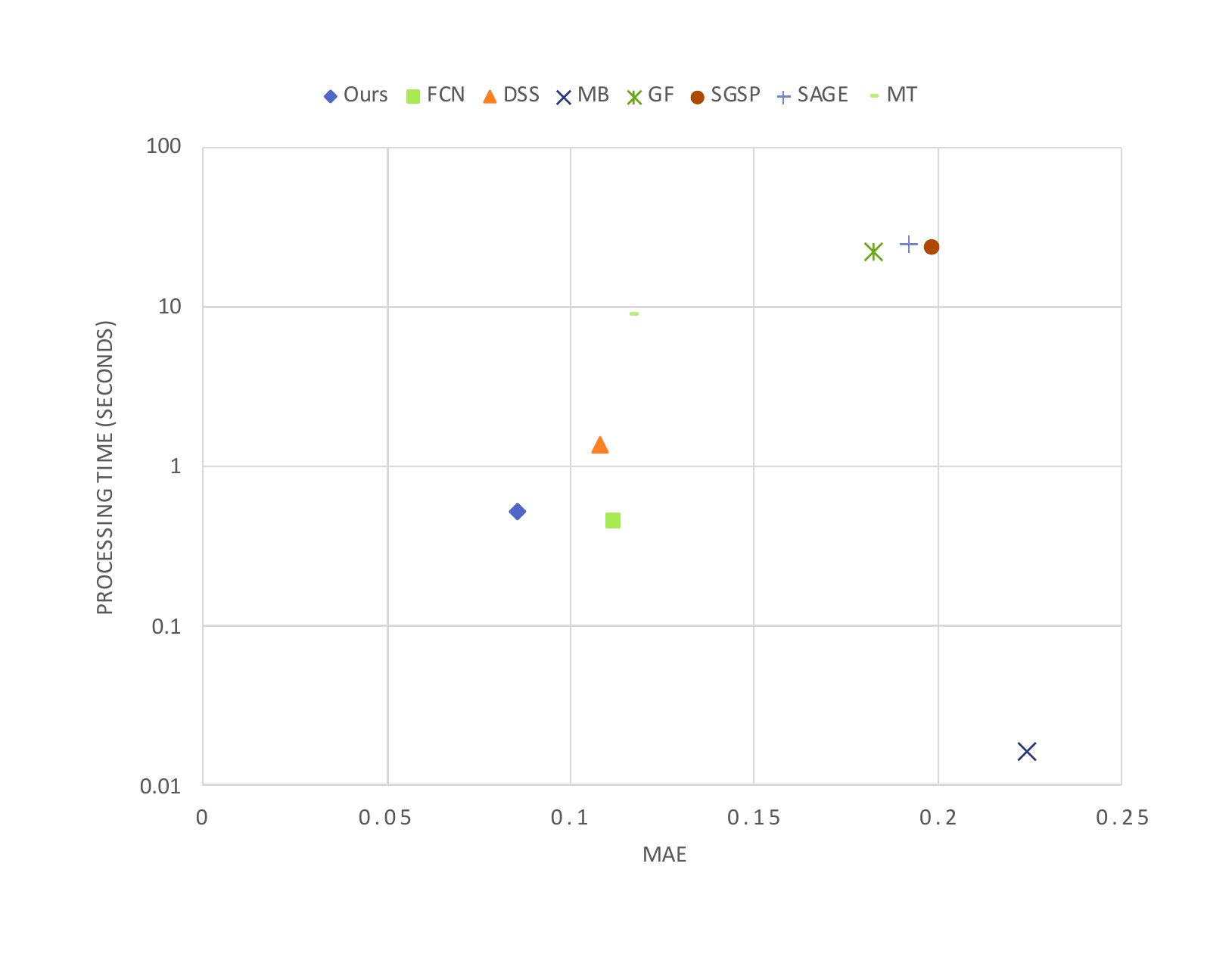}
  \end{minipage}
  
  \mbox{}\hfill (a) \hfill\mbox{}
  \mbox{}\hfill (b) \hfill\mbox{}
  \mbox{}\hfill (c) \hfill\mbox{}
  \caption{Quantitative comparison with state-of-the-art methods on DAVIS dataset (top) and FBMS dataset (bottom): (a) precision recall curves and (b) receiver operating characteristic curves calculated for binarized saliency predictions under varying thresholds in range [0, 255], (c) scatter diagram showing the average frame processing time and MAE values for different methods. }
  \label{fig:charts}
\end{figure*}

\begin{table}[t]
  \centering
  \resizebox{0.9\textwidth}{!}
  {
  \begin{tabular}{|c|c|c|c|c|c|c|c|c|c|}
  \hline
  Data Set                    
  & Metric     & DSS     & GF      & FCN     & MB      & SGSP    & MT & SAGE & Ours    \\ \hline
  & maxF & 0.72128 & 0.62104 & {\color[HTML]{32CB00} 0.74708} & 0.53342 & 0.6894  & {\color[HTML]{3166FF} 0.75288} & 0.5534 &  {\color[HTML]{FE0000} 0.8151}  \\ \cline{2-10}
  & avgF & {\color[HTML]{32CB00} 0.65464} & 0.51829 & {\color[HTML]{3166FF} 0.66715} & 0.41838 & 0.50776 & 0.63193 & 0.47085 & {\color[HTML]{FE0000} 0.72328} \\ \cline{2-10}
  & AUC  & 0.944   & 0.90792 & {\color[HTML]{32CB00} 0.95526} & 0.90228 & 0.94283 & {\color[HTML]{3166FF} 0.97326} & 0.93988 & {\color[HTML]{FE0000} 0.97483} \\ \cline{2-10}       
  \multirow{-4}{*}{DAVIS}   
  & MAE  & 0.08583 & 0.10023 & {\color[HTML]{3166FF} 0.0591}  & 0.17671 & 0.13833 & {\color[HTML]{32CB00} 0.07113} & 0.10277 & {\color[HTML]{FE0000} 0.05094} \\ \hline
  & maxF & {\color[HTML]{3166FF} 0.77823} & 0.58252 & {\color[HTML]{32CB00} 0.75159} & 0.54294 & 0.6494  & 0.7124 & 0.57559 & {\color[HTML]{FE0000} 0.82307} \\ \cline{2-10}
  & avgF & {\color[HTML]{FE0000} 0.74009} & 0.47498 & {\color[HTML]{32CB00} 0.68384} & 0.48426 & 0.49173 & 0.63285 & 0.4771 & {\color[HTML]{3166FF} 0.7389}  \\ \cline{2-10}
  & AUC  & 0.89815 & 0.82174 & {\color[HTML]{3166FF} 0.92002} & 0.83605 & 0.8364  & {\color[HTML]{32CB00} 0.90934} & 0.87128 & {\color[HTML]{FE0000} 0.96489} \\ \cline{2-10}
  \multirow{-4}{*}{FBMS} 
  & MAE  & {\color[HTML]{3166FF} 0.10808} & 0.18235 & {\color[HTML]{32CB00} 0.11161} & 0.22402 & 0.19863 & 0.11677 & 0.19216 & {\color[HTML]{FE0000} 0.0853} \\ \hline
  \end{tabular}
  }
  \caption{Quantitative comparison with state-of-the-art methods, using maximum and average F-measure (larger is better), AUC (larger is better) and MAE (smaller is better). The best three results are colored \color[HTML]{FE0000}\textbf{red}\color{black}, \color[HTML]{3166FF}\textbf{blue}\color{black}, and \color[HTML]{32CB00}\textbf{green}\color{black}, respectively.}
  \label{tab:comp_quantity}
\end{table}

The evaluation criteria described in Section~\ref{sec:criteria} are measured for different saliency prediction approaches and reported in Figure~\ref{fig:charts} and Table~\ref{tab:comp_quantity}. 

Precision recall and receiver operating characteristic curves are compared in Figure~\ref{fig:charts} along a scatter diagram showing mean absolute error and processing time for different methods. As it can be observed in \textbf{PR} and \textbf{ROC} curves, our method consistently outperforms all of the state-of-the-art static and dynamic saliency detection methods on both datasets. The value of mean absolute error (MAE) for the proposed method is also less than all other methods for the two datasets, as it is shown on the rightmost charts.

Quantitative comparison of four important metrics including maximum and average F-measure (\textbf{maxF} and \textbf{avgF}), area under ROC curve (\textbf{AUC}) and mean absolute error (\textbf{MAE}) is presented in Table~\ref{tab:comp_quantity}. The non-local deep saliency detection method improves maximum F-measure by 6.8\% and 7.1\% respectively on DAVIS and FBMS datasets over its baseline method while reducing MAE by 0.8\% and 2.6\% accordingly.

\subsection{Time comparison} 
One of the most important aspects to consider in saliency detection is the computational-efficiency of the algorithm as it is often used as a preprocessing step for machine vision applications. The proposed method increases processing time for each video frame, only 12\% compared to the baseline method which is considered one of the fastest video saliency detection methods. As shown in Figure~\ref{fig:charts} the only faster algorithm other than the baseline method in our experiments is Minimum Barrier Salient Object Detection (\textbf{MB}) \cite{MinimumBarrierSaliency2015} which produces much less accurate saliency maps using only static images. Compared to that algorithm, our method has 12.6\% and 13.9\% less MAE, respectively on DAVIS and FBMS datasets which is obviously worth the extra computational effort. A PC with a 3.4GHz Intel CPU and an NVIDIA Titan X GPU is used for performing the experiments.

\subsection{Ablation study}

\begin{table}[t]
  \centering
  \begin{adjustbox}{max width=0.5\textwidth}
  \begin{tabular}{c c c c }
  \toprule
  \textbf{After block \#}                    & \textbf{\# of NL blocks} & \textbf{MAE} & \textbf{Time} \\ \midrule
  \multirow{5}{*}{3}  & 1                            & 0.0519       & 0.812         \\  
                                                & 2                            & 0.0511       & 1.247         \\  
                                                & 3                            & 0.0508       & 1.553         \\  
                                                & 4                            & 0.0507       & 1.982         \\  
                                                & 5                            & 0.0507       & 2.353         \\ \midrule
  \multirow{5}{*}{4} & 1                            & 0.0526       & 0.561         \\  
                                                & 2                            & 0.0512       & 0.649         \\  
                                                & 3                            & 0.0509       & 0.732         \\  
                                                & 4                            & 0.0508       & 0.85          \\  
                                                & 5                            & 0.0508       & 0.958         \\ \midrule
  \multirow{5}{*}{5}  & 1                            & 0.0533       & 0.477         \\  
                                                & 2                            & 0.0514       & 0.494         \\  
                                                & 3                            & 0.0509       & 0.512         \\  
                                                & 4                            & 0.0509       & 0.531         \\  
                                                & 5                            & 0.0509       & 0.554         \\ \bottomrule
  \end{tabular}
  \end{adjustbox}
  \caption{Evaluation of the proposed architecure employing different number of non-local blocks placed after the last three convolution blocks, in terms of mean absolute error (MAE) and processing time.}
  \label{tab:ablation}
\end{table}

Non-local blocks are employed in the proposed architecture to exploit non-local appearance and motion information for detecting salient objects. Here, we study the effect of using these blocks after different convolutional layers varying the number of blocks used in each case.

Although employing non-local blocks after convolution layers with higher resolution has the advantage of capturing more detailed information over that of using them after layers having less resolution, the improvement is not substantial, according to the results reported in Table~\ref{tab:ablation}. Considering the computational overhead caused by placing non-local blocks after 3rd and 4th convolution blocks and the insignificant improvements compared to employing them after the last (5th) convolution block, the non-local blocks are used after the last convolution block in our network; Moreover, using more than three non-local blocks after the last convolution block does not seem to lessen the error according to the Table~\ref{tab:ablation}. From these observations, it has been decided to employ three non-local blocks after the fifth convolution block in our architecture.

\section{Conclusion}

In this paper, we investigated the application of non-local blocks in video salient object detection task and presented a deep CNN based framework for saliency detection, incorporating non-local operations to capture global appearance and motion information. The proposed method improves the results of its baseline method significantly and demonstrates the effectiveness of non-local operations for the task of salient object detection. Since non-local blocks can be easily plugged onto any CNN based network, the proposed approach can be extended to other deep CNN based saliency methods as well.

\bibliographystyle{plain}
\bibliography{./refs.bib}

\begin{thebibliography}{10}

\bibitem{tensorflow2015-whitepaper}
Mart\'{\i}n Abadi, Ashish Agarwal, Paul Barham, Eugene Brevdo, Zhifeng Chen,
  Craig Citro, Greg~S. Corrado, Andy Davis, Jeffrey Dean, Matthieu Devin,
  Sanjay Ghemawat, Ian Goodfellow, Andrew Harp, Geoffrey Irving, Michael Isard,
  Yangqing Jia, Rafal Jozefowicz, Lukasz Kaiser, Manjunath Kudlur, Josh
  Levenberg, Dandelion Man\'{e}, Rajat Monga, Sherry Moore, Derek Murray, Chris
  Olah, Mike Schuster, Jonathon Shlens, Benoit Steiner, Ilya Sutskever, Kunal
  Talwar, Paul Tucker, Vincent Vanhoucke, Vijay Vasudevan, Fernanda Vi\'{e}gas,
  Oriol Vinyals, Pete Warden, Martin Wattenberg, Martin Wicke, Yuan Yu, and
  Xiaoqiang Zheng.
\newblock {TensorFlow}: Large-scale machine learning on heterogeneous systems,
  2015.
\newblock Software available from tensorflow.org.

\bibitem{FrequencyTunedSRD2009}
Radhakrishna Achanta, Sheila~S. Hemami, Francisco~J. Estrada, and Sabine
  S{\"u}sstrunk.
\newblock Frequency-tuned salient region detection.
\newblock {\em 2009 IEEE Conference on Computer Vision and Pattern
  Recognition}, pages 1597--1604, 2009.

\bibitem{Preattentive2009}
Lawrence~G. Appelbaum and Anthony~M. Norcia.
\newblock Attentive and pre-attentive aspects of figural processing.
\newblock {\em Journal of Vision}, 9(11):18, 2009.

\bibitem{SpatioTemporalDynamicSaliency2018}
C.~Bak, A.~Kocak, E.~Erdem, and A.~Erdem.
\newblock Spatio-temporal saliency networks for dynamic saliency prediction.
\newblock {\em IEEE Transactions on Multimedia}, 20(7):1688--1698, July 2018.

\bibitem{SalientObjectDetectionBenchmark2015}
A.~Borji, M.~Cheng, H.~Jiang, and J.~Li.
\newblock Salient object detection: A benchmark.
\newblock {\em IEEE Transactions on Image Processing}, 24(12):5706--5722, Dec
  2015.

\bibitem{SalientObjectDetectionSurvey2014}
Ali Borji, Ming-Ming Cheng, Huaizu Jiang, and Jia Li.
\newblock Salient object detection: A survey.
\newblock {\em CoRR}, abs/1411.5878, 2014.

\bibitem{OSVOS2017}
Sergi Caelles, Kevis-Kokitsi Maninis, Jordi Pont-Tuset, Laura Leal-Taix{\'e},
  Daniel Cremers, and Luc~Van Gool.
\newblock One-shot video object segmentation.
\newblock {\em 2017 IEEE Conference on Computer Vision and Pattern Recognition
  (CVPR)}, pages 5320--5329, 2017.

\bibitem{GenericObjectnessSaliency2011}
Kai-Yueh Chang, Tyng-Luh Liu, Hwann-Tzong Chen, and Shang-Hong Lai.
\newblock Fusing generic objectness and visual saliency for salient object
  detection.
\newblock {\em 2011 International Conference on Computer Vision}, pages
  914--921, 2011.

\bibitem{VideoSaliencyCoherencyDiffusion2017}
C.~Chen, S.~Li, Y.~Wang, H.~Qin, and A.~Hao.
\newblock Video saliency detection via spatial-temporal fusion and low-rank
  coherency diffusion.
\newblock {\em IEEE Transactions on Image Processing}, 26(7):3156--3170, July
  2017.

\bibitem{SpectralGraph2018}
Jiazhong Chen, Jie Chen, Hefei Ling, Hua Cao, Weiping Sun, Yebin Fan, and
  Weimin Wu.
\newblock Salient object detection via spectral graph weighted low rank matrix
  recovery.
\newblock {\em Journal of Visual Communication and Image Representation},
  50:270 -- 279, 2018.

\bibitem{GlobalContrastSaliency2011}
Ming-Ming Cheng, Guo-Xin Zhang, Niloy~Jyoti Mitra, Xiaolei Huang, and Shi-Min
  Hu.
\newblock Global contrast based salient region detection.
\newblock {\em CVPR 2011}, pages 409--416, 2011.

\bibitem{VisualSaliencyReview2018}
Runmin Cong, Jianjun Lei, Huazhu Fu, Ming-Ming Cheng, Weisi Lin, and Qingming
  Huang.
\newblock Review of visual saliency detection with comprehensive information.
\newblock {\em CoRR}, abs/1803.03391, 2018.

\bibitem{ImageNet2009}
Jia Deng, Wei Dong, Richard Socher, Li-Jia Li, Kai Li, and Li~Fei-Fei.
\newblock Imagenet: A large-scale hierarchical image database.
\newblock {\em 2009 IEEE Conference on Computer Vision and Pattern
  Recognition}, pages 248--255, 2009.

\bibitem{VideoSaliency3dCNN2018}
Guanqun Ding and Yuming Fang.
\newblock Video saliency detection by 3d convolutional neural networks.
\newblock In Guangtao Zhai, Jun Zhou, and Xiaokang Yang, editors, {\em Digital
  TV and Wireless Multimedia Communication}, pages 245--254, Singapore, 2018.
  Springer Singapore.

\bibitem{VideoSaliencyUncertaintyWeighting2014}
Y.~Fang, Z.~Wang, W.~Lin, and Z.~Fang.
\newblock Video saliency incorporating spatiotemporal cues and uncertainty
  weighting.
\newblock {\em IEEE Transactions on Image Processing}, 23(9):3910--3921, Sept
  2014.

\bibitem{SaliencyGuidedRegionProposal2017}
A.~Fattal, M.~Karg, C.~Scharfenberger, and J.~Adamy.
\newblock Saliency-guided region proposal network for cnn based object
  detection.
\newblock In {\em 2017 IEEE 20th International Conference on Intelligent
  Transportation Systems (ITSC)}, pages 1--8, Oct 2017.

\bibitem{VOCUS2006}
Simone Frintrop.
\newblock {\em VOCUS: A Visual Attention System for Object Detection and
  Goal-Directed Search (Lecture Notes in Computer Science / Lecture Notes in
  Artificial Intelligence)}.
\newblock Springer-Verlag, Berlin, Heidelberg, 2006.

\bibitem{MultipleForegroundSegmentation2017}
H.~Fu, D.~Xu, and S.~Lin.
\newblock Object-based multiple foreground segmentation in rgbd video.
\newblock {\em IEEE Transactions on Image Processing}, 26(3):1418--1427, March
  2017.

\bibitem{FastRCNN2015}
Ross~B. Girshick.
\newblock Fast r-cnn.
\newblock {\em 2015 IEEE International Conference on Computer Vision (ICCV)},
  pages 1440--1448, 2015.

\bibitem{ContextAwareSaliency2010}
Stas Goferman, Lihi Zelnik-Manor, and Ayellet Tal.
\newblock Context-aware saliency detection.
\newblock {\em 2010 IEEE Computer Society Conference on Computer Vision and
  Pattern Recognition}, pages 2376--2383, 2010.

\bibitem{VideoSaliencyObjectProposals2018}
F.~Guo, W.~Wang, J.~Shen, L.~Shao, J.~Yang, D.~Tao, and Y.~Y. Tang.
\newblock Video saliency detection using object proposals.
\newblock {\em IEEE Transactions on Cybernetics}, pages 1--12, 2018.

\bibitem{ObjectDetectionSelectiveAttention2014}
Mingwei Guo, Yuzhou Zhao, Chenbin Zhang, and Zonghai Chen.
\newblock Fast object detection based on selective visual attention.
\newblock {\em Neurocomputing}, 144:184 -- 197, 2014.

\bibitem{MaskRCNN2017}
Kaiming He, Georgia Gkioxari, Piotr Doll{\'a}r, and Ross~B. Girshick.
\newblock Mask r-cnn.
\newblock {\em 2017 IEEE International Conference on Computer Vision (ICCV)},
  pages 2980--2988, 2017.

\bibitem{SaliencyShortConnections2018}
Q.~Hou, M.~Cheng, X.~Hu, A.~Borji, Z.~Tu, and P.~H.~S. Torr.
\newblock Deeply supervised salient object detection with short connections.
\newblock {\em IEEE Transactions on Pattern Analysis and Machine Intelligence},
  pages 1--1, 2018.

\bibitem{VideoSaliencyDynamicAttention2013}
Sheng hua Zhong, Yan Liu, Feifei Ren, Jinghuan Zhang, and Tongwei Ren.
\newblock Video saliency detection via dynamic consistent spatio-temporal
  attention modelling.
\newblock In {\em AAAI}, 2013.

\bibitem{SaliencySearchItti2000}
Laurent Itti and Christof Koch.
\newblock A saliency-based search mechanism for overt and covert shifts of
  visual attention.
\newblock {\em Vision Research}, 40:1489--1506, 2000.

\bibitem{SaliencyAttentionSceneAnalysis1998}
Laurent Itti, Christof Koch, and Ernst Niebur.
\newblock A model of saliency-based visual attention for rapid scene analysis.
\newblock {\em IEEE Trans. Pattern Anal. Mach. Intell.}, 20:1254--1259, 1998.

\bibitem{VideoSummarizationAttention2017}
Zhong Ji, Kailin Xiong, Yanwei Pang, and Xuelong Li.
\newblock Video summarization with attention-based encoder-decoder networks.
\newblock {\em CoRR}, abs/1708.09545, 2017.

\bibitem{ObjectLevelSaliency2013}
Yangqing Jia and Mei Han.
\newblock Category-independent object-level saliency detection.
\newblock {\em 2013 IEEE International Conference on Computer Vision}, pages
  1761--1768, 2013.

\bibitem{ObjectToMotionLCNN2017}
Lai Jiang, Mai Xu, and Zulin Wang.
\newblock Predicting video saliency with object-to-motion cnn and two-layer
  convolutional lstm.
\newblock {\em CoRR}, abs/1709.06316, 2017.

\bibitem{Difnet2018}
Peng Jiang, Fanglin Gu, Changhe Tu, and Baoquan Chen.
\newblock Difnet: Semantic segmentation by diffusion networks.
\newblock {\em CoRR}, abs/1805.08015, 2018.

\bibitem{SalientRegionDetection2015}
Rajkumar Kannan, Gheorghita Ghinea, and Sridhar Swaminathan.
\newblock Discovering salient objects from videos using spatiotemporal salient
  region detection.
\newblock {\em Signal Processing: Image Communication}, 36:154 -- 178, 2015.

\bibitem{CenterSurroundedSaliency2011}
Dominik~A. Klein and Simone Frintrop.
\newblock Center-surround divergence of feature statistics for salient object
  detection.
\newblock {\em 2011 International Conference on Computer Vision}, pages
  2214--2219, 2011.

\bibitem{RegionBasedSaliency2017}
Trung-Nghia Le and Akihiro Sugimoto.
\newblock Region-based multiscale spatiotemporal saliency for video.
\newblock {\em CoRR}, abs/1708.01589, 2017.

\bibitem{VideoSaliencyDeepFeatures2018}
Trung-Nghia Le and Akihiro Sugimoto.
\newblock Video salient object detection using spatiotemporal deep features.
\newblock {\em IEEE Transactions on Image Processing}, 27:5002--5015, 2018.

\bibitem{UniversalSaliencyFramework2016}
Jianjun Lei, Bingren Wang, Yuming Fang, Weisi Lin, Patrick~Le Callet, Nam Ling,
  and Chunping Hou.
\newblock A universal framework for salient object detection.
\newblock {\em IEEE Transactions on Multimedia}, 18:1783--1795, 2016.

\bibitem{InstanceLevelSaliency2017}
Guanbin Li, Yuan Xie, Liang Lin, and Yizhou Yu.
\newblock Instance-level salient object segmentation.
\newblock {\em 2017 IEEE Conference on Computer Vision and Pattern Recognition
  (CVPR)}, pages 247--256, 2017.

\bibitem{FlowGuidedSaliency2018}
Guanbin Li, Yuan Xie, Tianhao Wei, Keze Wang, and Liang Lin.
\newblock Flow guided recurrent neural encoder for video salient object
  detection.
\newblock In {\em The IEEE Conference on Computer Vision and Pattern
  Recognition (CVPR)}, June 2018.

\bibitem{VisualSaliencyMultiscale2015}
Guanbin Li and Yizhou Yu.
\newblock Visual saliency based on multiscale deep features.
\newblock {\em 2015 IEEE Conference on Computer Vision and Pattern Recognition
  (CVPR)}, pages 5455--5463, 2015.

\bibitem{DeepContrastSalientObject2016}
Guanbin Li and Yizhou Yu.
\newblock Deep contrast learning for salient object detection.
\newblock {\em CoRR}, abs/1603.01976, 2016.

\bibitem{MultiTaskDeepSaliency2016}
X.~Li, L.~Zhao, L.~Wei, M.~Yang, F.~Wu, Y.~Zhuang, H.~Ling, and J.~Wang.
\newblock Deepsaliency: Multi-task deep neural network model for salient object
  detection.
\newblock {\em IEEE Transactions on Image Processing}, 25(8):3919--3930, Aug
  2016.

\bibitem{COCO2014}
Tsung-Yi Lin, Michael Maire, Serge~J. Belongie, James Hays, Pietro Perona, Deva
  Ramanan, Piotr Doll{\'a}r, and C.~Lawrence Zitnick.
\newblock Microsoft coco: Common objects in context.
\newblock In {\em ECCV}, 2014.

\bibitem{NonLocalRecurrentImageRestoration2018}
Ding Liu, Bihan Wen, Yuchen Fan, Chen~Change Loy, and Thomas~S. Huang.
\newblock Non-local recurrent network for image restoration.
\newblock {\em CoRR}, abs/1806.02919, 2018.

\bibitem{HirerchicalSaliency2016}
N.~Liu and J.~Han.
\newblock Dhsnet: Deep hierarchical saliency network for salient object
  detection.
\newblock In {\em 2016 IEEE Conference on Computer Vision and Pattern
  Recognition (CVPR)}, pages 678--686, June 2016.

\bibitem{VideoAttention2008}
Tie Liu, Nanning Zheng, Wei Ding, and Zejian Yuan.
\newblock Video attention: Learning to detect a salient object sequence.
\newblock {\em 2008 19th International Conference on Pattern Recognition},
  pages 1--4, 2008.

\bibitem{SSD2016}
Wei Liu, Dragomir Anguelov, Dumitru Erhan, Christian Szegedy, Scott~E. Reed,
  Cheng-Yang Fu, and Alexander~C. Berg.
\newblock Ssd: Single shot multibox detector.
\newblock In {\em ECCV}, 2016.

\bibitem{UnconstrainedSuperpixelSaliency2017}
Z.~Yong Liu, Junhao Li, Linwei Ye, Guangling Sun, and Liquan Shen.
\newblock Saliency detection for unconstrained videos using superpixel-level
  graph and spatiotemporal propagation.
\newblock {\em IEEE Transactions on Circuits and Systems for Video Technology},
  27:2527--2542, 2017.

\bibitem{UnconstrainedSalientObject2017}
Mahyar Najibi, Fan Yang, Qiaosong Wang, and Robinson Piramuthu.
\newblock Towards the success rate of one: Real-time unconstrained salient
  object detection.
\newblock {\em CoRR}, abs/1708.00079, 2017.

\bibitem{IntegratedAttentionOptimizingDetection2006}
V.~Navalpakkam and L.~Itti.
\newblock An integrated model of top-down and bottom-up attention for
  optimizing detection speed.
\newblock In {\em 2006 IEEE Computer Society Conference on Computer Vision and
  Pattern Recognition (CVPR'06)}, volume~2, pages 2049--2056, June 2006.

\bibitem{FBMS2014}
P.~Ochs, J.~Malik, and T.~Brox.
\newblock Segmentation of moving objects by long term video analysis.
\newblock {\em IEEE Transactions on Pattern Analysis and Machine Intelligence},
  36(6):1187--1200, June 2014.

\bibitem{AttentionUnet2018}
Ozan Oktay, Jo~Schlemper, Lo{\"i}c~Le Folgoc, Matthew C.~H. Lee, Mattias~P.
  Heinrich, Kazunari Misawa, Kensaku Mori, Steven~G. McDonagh, Nils~Y.
  Hammerla, Bernhard Kainz, Ben Glocker, and Daniel Rueckert.
\newblock Attention u-net: Learning where to look for the pancreas.
\newblock {\em CoRR}, abs/1804.03999, 2018.

\bibitem{DAVIS2016}
F.~Perazzi, J.~Pont-Tuset, B.~McWilliams, L.~V. Gool, M.~Gross, and
  A.~Sorkine-Hornung.
\newblock A benchmark dataset and evaluation methodology for video object
  segmentation.
\newblock In {\em 2016 IEEE Conference on Computer Vision and Pattern
  Recognition (CVPR)}, pages 724--732, June 2016.

\bibitem{SaliencyFilters2012}
Federico Perazzi, Philipp Kr{\"a}henb{\"u}hl, Yael Pritch, and Alexander
  Sorkine-Hornung.
\newblock Saliency filters: Contrast based filtering for salient region
  detection.
\newblock {\em 2012 IEEE Conference on Computer Vision and Pattern
  Recognition}, pages 733--740, 2012.

\bibitem{SegmentingSalientObject2010}
Esa Rahtu, Juho Kannala, Mikko Salo, and Janne Heikkil{\"a}.
\newblock Segmenting salient objects from images and videos.
\newblock In {\em ECCV}, 2010.

\bibitem{YOLO2016}
Joseph Redmon, Santosh~Kumar Divvala, Ross~B. Girshick, and Ali Farhadi.
\newblock You only look once: Unified, real-time object detection.
\newblock {\em 2016 IEEE Conference on Computer Vision and Pattern Recognition
  (CVPR)}, pages 779--788, 2016.

\bibitem{UNET2015}
Olaf Ronneberger, Philipp Fischer, and Thomas Brox.
\newblock U-net: Convolutional networks for biomedical image segmentation.
\newblock In {\em MICCAI}, 2015.

\bibitem{VeryDeepCNN2014}
Karen Simonyan and Andrew Zisserman.
\newblock Very deep convolutional networks for large-scale image recognition.
\newblock {\em CoRR}, abs/1409.1556, 2014.

\bibitem{VideoHashingDBN2016}
Jiande Sun, Xiaocui Liu, Wenbo Wan, Jing Li, Dong Zhao, and Huaxiang Zhang.
\newblock Video hashing based on appearance and attention features fusion via
  dbn.
\newblock {\em Neurocomputing}, 213:84--94, 2016.

\bibitem{MultiScaleConvLSTMSaliency2018}
Yi~Tang, Wenbin Zou, Zhi Jin, and Xia Li.
\newblock Multi-scale spatiotemporal conv-lstm network for video saliency
  detection.
\newblock In {\em Proceedings of the 2018 ACM on International Conference on
  Multimedia Retrieval}, ICMR '18, pages 362--369, New York, NY, USA, 2018.
  ACM.

\bibitem{SalientObjectRecurrentFCNN2018}
L.~Wang, L.~Wang, H.~Lu, P.~Zhang, and X.~Ruan.
\newblock Salient object detection with recurrent fully convolutional networks.
\newblock {\em IEEE Transactions on Pattern Analysis and Machine Intelligence},
  pages 1--1, 2018.

\bibitem{DeepSaliencyLocalEstimation2015}
Lijun Wang, Huchuan Lu, Xiang Ruan, and Ming-Hsuan Yang.
\newblock Deep networks for saliency detection via local estimation and global
  search.
\newblock {\em 2015 IEEE Conference on Computer Vision and Pattern Recognition
  (CVPR)}, pages 3183--3192, 2015.

\bibitem{DeepAttentionPrediction2018}
W.~Wang and J.~Shen.
\newblock Deep visual attention prediction.
\newblock {\em IEEE Transactions on Image Processing}, 27(5):2368--2378, May
  2018.

\bibitem{ConsistentSaliencyGradientFlow2015}
W.~Wang, J.~Shen, and L.~Shao.
\newblock Consistent video saliency using local gradient flow optimization and
  global refinement.
\newblock {\em IEEE Transactions on Image Processing}, 24(11):4185--4196, Nov
  2015.

\bibitem{VideoSaliancyFCN2018}
W.~Wang, J.~Shen, and L.~Shao.
\newblock Video salient object detection via fully convolutional networks.
\newblock {\em IEEE Transactions on Image Processing}, 27(1):38--49, Jan 2018.

\bibitem{SaliencyAwareObjectSegmentation2015}
Wenguan Wang, Jianbing Shen, and F.~Porikli.
\newblock Saliency-aware geodesic video object segmentation.
\newblock In {\em 2015 IEEE Conference on Computer Vision and Pattern
  Recognition (CVPR)}, pages 3395--3402, June 2015.

\bibitem{SaliencyAwareObjectSegmentation2018}
Wenguan Wang, Jianbing Shen, Ruigang Yang, and Fatih~Murat Porikli.
\newblock Saliency-aware video object segmentation.
\newblock {\em IEEE Transactions on Pattern Analysis and Machine Intelligence},
  40:20--33, 2018.

\bibitem{NonLocal2018}
Xiaolong Wang, Ross Girshick, Abhinav Gupta, and Kaiming He.
\newblock Non-local neural networks.
\newblock {\em CVPR}, 2018.

\bibitem{SaliencyAwareActionRecognition2017}
Xuanhan Wang, Lianli Gao, Jingkuan Song, and Heng~Tao Shen.
\newblock Beyond frame-level cnn: Saliency-aware 3-d cnn with lstm for video
  action recognition.
\newblock {\em IEEE Signal Processing Letters}, 24:510--514, 2017.

\bibitem{FeatureHybridFrameworkSaliency2018}
Zheng Wang, Jinchang Ren, Dong Zhang, Meijun Sun, and Jianmin Jiang.
\newblock A deep-learning based feature hybrid framework for spatiotemporal
  saliency detection inside videos.
\newblock {\em Neurocomputing}, 287:68--83, 2018.

\bibitem{SpatiotemporalAttention2006}
Yun Zhai and Mubarak Shah.
\newblock Visual attention detection in video sequences using spatiotemporal
  cues.
\newblock In {\em ACM Multimedia}, 2006.

\bibitem{MinimumBarrierSaliency2015}
Jianming Zhang, Stan Sclaroff, Zhe~L. Lin, Xiaohui Shen, Brian~L. Price, and
  Radom{\'i}r Mech.
\newblock Minimum barrier salient object detection at 80 fps.
\newblock {\em 2015 IEEE International Conference on Computer Vision (ICCV)},
  pages 1404--1412, 2015.

\bibitem{UnconstrainedSaliencyProposal2016}
Jianming Zhang, Stan Sclaroff, Zhe~L. Lin, Xiaohui Shen, Brian~L. Price, and
  Radom{\'i}r Mech.
\newblock Unconstrained salient object detection via proposal subset
  optimization.
\newblock {\em 2016 IEEE Conference on Computer Vision and Pattern Recognition
  (CVPR)}, pages 5733--5742, 2016.

\bibitem{TimeMappingSaliency2014}
F.~Zhou, S.~B. Kang, and M.~F. Cohen.
\newblock Time-mapping using space-time saliency.
\newblock In {\em 2014 IEEE Conference on Computer Vision and Pattern
  Recognition}, pages 3358--3365, June 2014.

\bibitem{BaggingBasedPrediction2018}
Xiaofei Zhou, Zhi Liu, Kai Li, and Guangling Sun.
\newblock Video saliency detection via bagging-based prediction and
  spatiotemporal propagation.
\newblock {\em Journal of Visual Communication and Image Representation},
  51:131 -- 143, 2018.

\end{thebibliography}

\end{document}